%% file: lrec-coling2024.tex
\documentclass[10pt, a4paper]{article}

\usepackage[table,xcdraw]{xcolor} 
\usepackage{lrec-coling2024} 
\usepackage{comment}
\usepackage{booktabs}
\usepackage{enumitem}
\usepackage{adjustbox}
\usepackage{multirow}
\usepackage{hyperref}
\usepackage{svg}
\usepackage{pgfplots} 
\usepackage{pgfplots, pgfplotstable}
\usetikzlibrary{patterns}
\usepackage{subcaption}
\usepackage{bold-extra}
\usepackage{tabularx}

\newcommand{\shaderow}{\rowcolor{lightgray!30}[0pt][0pt]}
\newcommand\uline[1]{\underline{\smash{#1}}}
\newcommand{\iathree}{$\mbox{(IA)}^{\scriptsize{\mbox{3}}}$ }
\newcounter{rownumber}[table] 
\title{ILLUMINER: Instruction-tuned Large Language Models as Few-shot Intent Classifier and Slot Filler}

\name{Paramita Mirza$^{\ast}$, Viju Sudhi$^{\dagger}$, Soumya Ranjan Sahoo$^{\dagger}$, \\ {\bf \large Sinchana Ramakanth Bhat$^{\dagger}$ }}

\address{$^{\ast}$Fraunhofer IIS, $^{\dagger}$Fraunhofer IAIS \\
         paramita.paramita@iis.fraunhofer.de, \\ \{viju.sudhi, soumya.ranjan.sahoo, sinchana.ramakanth.bhat\}@iais.fraunhofer.de\\
         }

\abstract{
State-of-the-art intent classification (IC) and slot filling (SF) methods often rely on data-intensive deep learning models, limiting their practicality for industry applications.
Large language models on the other hand, particularly instruction-tuned models (Instruct-LLMs), exhibit remarkable zero-shot performance across various natural language tasks. This study 
evaluates
Instruct-LLMs on popular benchmark datasets for IC and SF, emphasizing their capacity to learn from fewer examples. We introduce ILLUMINER, an approach framing IC and SF as language generation tasks for Instruct-LLMs,
with a more efficient SF-prompting method compared to prior work.
A comprehensive comparison with multiple baselines shows
that our approach, using the FLAN-T5 11B model, outperforms the state-of-the-art joint IC+SF method and in-context learning with GPT3.5 (175B), 
particularly in slot filling by 11.1--32.2 percentage points. 
Additionally, our in-depth ablation study demonstrates that parameter-efficient fine-tuning requires less than 6\% of training data to yield comparable performance with traditional full-weight fine-tuning. 
 \\ \newline  }

\begin{document}

\maketitleabstract

\begin{figure*}[t]
    \begin{center}
    \begin{adjustbox}{width=1.0\textwidth}
    \includegraphics[trim=20 168 20 95,clip,width=\textwidth]{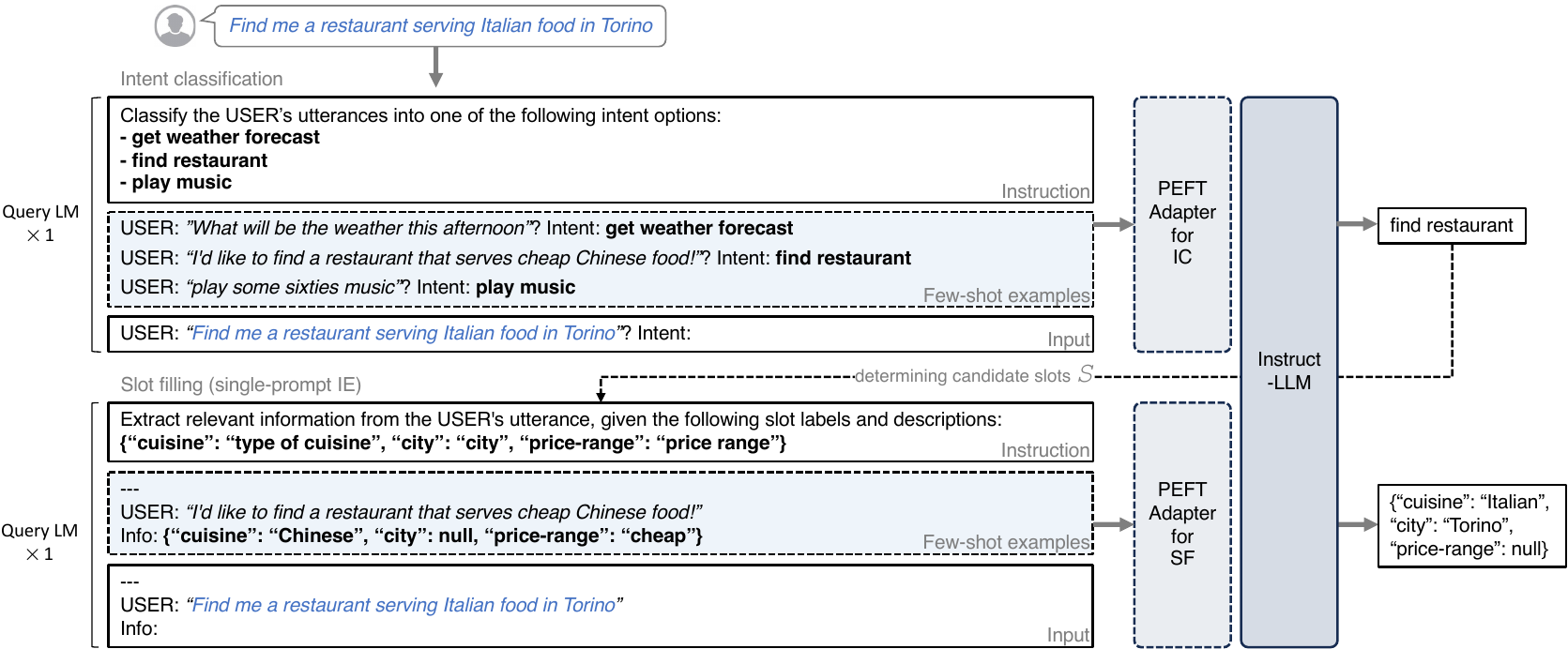}
    \end{adjustbox}
    \end{center}
    \caption{An example of our prompting methods for intent classification and slot filling, for a given user utterance \emph{``Find me a restaurant serving Italian food in Torino''}. Compared to prior work (Fig.~\ref{fig:SF-inverse-prompting}), we only need 
    a single inference
    for slot filling.}
    \label{fig:IC-SF-prompting}
\end{figure*}

\begin{figure}[t]
    \begin{center}
    \begin{adjustbox}{width=0.48\textwidth}
    \includegraphics[trim=80 302 330 97,clip,width=\textwidth]{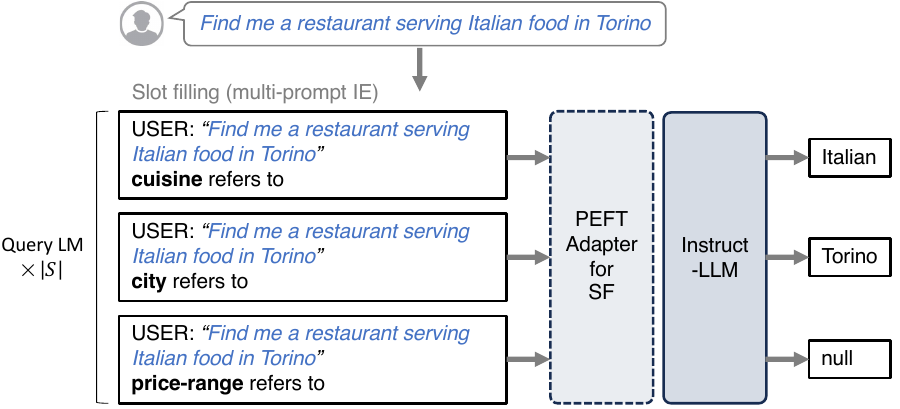}
    \end{adjustbox}
    \end{center}
    \caption{Multi-prompt IE for slot filling \citep{hou_inverse_2022} requiring $|S|$ inferences for $|S|$ slot types.}
    \label{fig:SF-inverse-prompting}
\end{figure}

\section{Introduction}
\input{introduction}

\section{Related Work}
\input{related}

\section{Methodology}
\input{methodology}

\input{experiments}


\nocite{*}
\newpage
\section{Bibliographical References}\label{sec:reference}

\bibliographystyle{lrec-coling2024-natbib}
\bibliography{lrec-coling2024}

\section{Language Resource References}
\label{lr:ref}
\bibliographystylelanguageresource{lrec-coling2024-natbib}
\bibliographylanguageresource{languageresource}

\end{document}

%% file: introduction.tex
Intent classification (IC) and slot filling (SF) are foundational tasks in natural language understanding (NLU) within task-oriented dialogue (TOD) systems,
which enable users to 
interact in natural language, facilitating various actions such as reserving a restaurant or seeking customer support. 
For instance, given a user utterance \emph{``Find me a restaurant serving Italian food in Torino''}, IC discerns the user's intent as \emph{find restaurant}, while SF aims for extracting slot type--value pairs \{(\emph{cuisine, `Italian'}), (\emph{city, `Torino'})\} from the utterance. This information is crucial for 
generating appropriate system responses.
Furthermore, efficiently and reliably solving these tasks with low latency is vital for the widespread deployment of TOD systems. 
Although deep learning models have excelled in supervised learning approaches \citep{gupta_simple_2019,chen_bert_2019,han_bi-directional_2022}, 
their reliance on large-scale annotated data constrains their practical use in real-world industrial scenarios.

Large language models (LLMs), especially those fine-tuned with instructions (Instruct-LLMs), have been touted as effective zero-shot learners \cite{flant5}. Instruction tuning empowers these models to interpret and execute user instructions effectively, thereby controlling their behavior \cite{instruct-llm-survey}. 
Unlike supervised fine-tuning, which relies on input examples and their corresponding outputs, instruction tuning augments input--output examples with \emph{instructions} as high-level task descriptions (depicted in Figure~\ref{fig:IC-SF-prompting}). This allows instruction-tuned models to generalize more readily to new tasks or domains.
When combined with \emph{in-context learning} \citep{brown_language_2020}, 
where the model is exposed to input--output examples within the \emph{prompt},
LLM-prompting methods offer substantial benefits over traditional supervised approaches in terms of reduced labeled data requirements. 

In-context learning (ICL), or \emph{few-shot learning}, offers language models a chance to learn from examples, but models' context size often limits the number of examples.
Processing $k$ training examples for $k$-shot ICL also increases inference time $k$ times as the prompt size grows \citep{liu_few-shot_2022}.
While fine-tuning LLMs with more examples from downstream datasets yields substantial performance gains compared to using them out-of-the-box \citep{su_multi-task_2022,xie_unifiedskg_2022}, full-weight fine-tuning on consumer hardware is impractical and risks \emph{catastrophic forgetting} \citep{goodfellow_empirical_2015}, particularly when the downstream dataset is small and lacks diversity. \emph{Parameter-efficient fine-tuning} (PEFT, e.g. \citealt{peft-lora,liu_few-shot_2022}) alleviates these issues by allowing fine-tuning of a small number of additional parameters while freezing most LLM parameters, significantly reducing computational and storage costs while retaining the LLMs' prior, generalized knowledge.

\paragraph{Approach and Contributions.} In this work, we introduce our approach {\bf ILLUMINER}\footnote{\url{https://github.com/OpenGPTX/illuminer}}, {\bf I}nstruction-tuned {\bf L}arge {\bf L}ang{\bf U}age {\bf M}odels as {\bf IN}tent Classifier and Slot Fill{\bf ER}.
We formulate IC and SF as language generation tasks, as exemplified in Figure~\ref{fig:IC-SF-prompting}.
For intent classification, we list possible intent labels to choose from in the instruction, and expect the Instruct-LLM to generate the appropriate label reflecting the intent of the input utterance.
As opposed to prior work on slot filling with multiple prompts \citep{hou_inverse_2022} illustrated in Figure~\ref{fig:SF-inverse-prompting}, we adopt a \emph{single-prompt Information Extraction (IE)} approach, requiring a single query per utterance for the Instruct-LLM to generate slot type--value pairs.
We explore the performance of Instruct-LLMs further fine-tuned with task-specific instructions and domain-specific examples using PEFT approaches like \textit{Low-Rank Adaptation} (LoRA, \citealt{peft-lora}) and \textit{Infused Adapter by Inhibiting and Amplifying Inner Activations} \iathree by \citet{liu_few-shot_2022}. 

\vspace{2pt}
The salient contributions of our work are:
\begin{itemize}[leftmargin=*,topsep=0pt,itemsep=0pt]
    \item Exploratory analysis of prompt engineering for IC and SF, and a much more efficient SF-prompting method compared to existing techniques (e.g., \citealt{hou_inverse_2022,li_generative_2023}).
    \item Comprehensive comparative analysis of several Instruct-LLMs on popular benchmark datasets for IC and SF, including SNIPS \citeplanguageresource{coucke2018snips}, MASSIVE \citeplanguageresource{fitzgerald2022massive} and MultiWoz \citeplanguageresource{budzianowski2018multiwoz}, in various settings: zero-shot learning, few-shot learning, and PEFT. We demonstrate that ILLUMINER (with PEFT) outperforms state-of-the-art baselines, particularly in SF, given less than 6\% of training data.
    \item Extensive ablation study examining the impact of Instruct-LLMs, different PEFT techniques, model size, number of examples for fine-tuning, and label exposure in instructions, as well as generalization across datasets.
\end{itemize}


%% file: related.tex
\paragraph{PLMs for IC and SF.} The rise of pre-trained language models (PLMs) like BERT \citep{devlin_bert_2019} and RoBERTa \citep{liu_roberta_2019} has spurred extensive research in utilizing contextual embeddings for sequence classification and labeling, notably in the joint task of IC and SF \citep{gupta_simple_2019,chen_bert_2019,han_bi-directional_2022}, as has been well-documented by \citet{weld_survey_2023}. A joint IC+SF model 
offers the advantage of training/fine-tuning a single model while capitalizing on label correlations between intents and slots. However, it demands a large annotated corpus \citep{weld_survey_2023}, rendering its application impractical in real-world scenarios.

\vspace{2pt}
Addressing the few-shot scenarios of IC and SF, existing work explores PLMs from three main perspectives:
\emph{(1) task-adaptive fine-tuning} \citep{zhang_few-shot_2021,yu_few-shot_2021,ma_frustratingly_2021,hou_learning_2021}, 
\emph{(2) data augmentation} \citep{rosenbaum_linguist_2022,lin_selective_2023}, and 
\emph{(3) prompt-based learning} \citep{hou_inverse_2022,parikh_exploring_2023}. 
Our work aligns with prompt-based learning, a crucial consideration in low-resource scenarios where fine-tuning large PLMs (LLMs) is not feasible \citep{radford_language_2019,schick_exploiting_2021}.
Limited studies focus on prompt-based learning for IC or SF. \citet{parikh_exploring_2023} explore different zero- and few-shot methods for IC, including LLM prompting and parameter-efficient fine-tuning. \citet{hou_inverse_2022} introduce a multi-prompt method for SF (Fig.~\ref{fig:SF-inverse-prompting}), accelerating inference compared to the classic prompting that requires inference for every n-gram word span \citep{cui_template-based_2021}. Yet, a comprehensive evaluation of 
prompt-based learning for both IC and SF jointly is lacking.

\vspace{2pt}
While previous studies \citep{wang_slot_2022,hu_-context_2022,gupta_show_2022,hudecek_are_2023} explore prompt-based learning for the dialog state tracking (DST) task, they mostly focus on the dialog-level performance, making it difficult to analyze the performance on single dialog turns and on the specific sub-tasks of DST: IC and SF. Bridging this gap between the two lines of work (IC+SF vs DST) and demonstrating our approach's generality, we also evaluate ILLUMINER on MultiWoz \citep{budzianowski2018multiwoz}, a prominent benchmark dataset for DST.

\paragraph{Parameter-efficient Fine-tuning (PEFT).} 

Fine tuning large language models is often a compute-intensive task, demanding immoderate time, cost and monitoring resources. However, recent advances in 
PEFT 
address these challenges by learning considerably fewer LLM parameters 
while achieving comparable performance to models with fully fine-tuned weights.
Notable PEFT techniques include adapter tuning \citep{houlsby_parameter-efficient_2019}, prefix tuning \citep{li_prefix-tuning_2021}, prompt tuning \citep{lester_power_2021}, LoRA \citep{peft-lora} and \iathree \citep{liu2022fewshot}. 


\vspace{2pt}
In the realm of task-oriented dialogue systems, 
PEFT has been explored for IC, SF and response generation.
\citet{hung2022dstod} train adapters for individual domains, demonstrating their composition for multi-domain specialization. \citet{Wang_2022} use adapter tuning with a copy network to prevent catastrophic forgetting and ensures entity consistency in dialogue flow. \citet{fuisz2022improved} employ lightweight adapters on QA-tuned PLMs for SF treated as a Question Answering task.
\citet{li_generative_2023} explore prefix tuning for cross-domain SF, while \citet{chang2023prompting} demonstrate the use of prompt tuning for IC and SF with speech models. \citet{kwon-etal-2023-sidlr} show that multilingual mT0 models fine-tuned with LoRA outperform baselines on IC and SF sub-tasks for low-resource languages. Additionally, \citet{parikh_exploring_2023} demonstrate how FLAN-T5, fine-tuned with \iathree adapters, outperforms larger language models like GPT-3 in IC. 


\vspace{2pt}
Existing work examined different PEFT techniques individually on benchmark datasets. To the best of our knowledge, our work is the first to compare and contrast different PEFT techniques for fine-tuning LLMs for IC and SF, including an exploration of cross-dataset generalization offered by each technique.

\paragraph{Instruction Tuning (IT).}
Firstly introduced by 
\citet{flant5}, 
IT explores language models' cross-task generalization through supervised fine-tuning with task-specific instructions and desired output \citep{instruct-llm-survey}.
Aligning the next-word prediction with user instructions enhances control and predictability, gaining traction with models like InstructGPT \cite{ouyang_training_2022} and FLAN-T5 \cite{flant5}, which oftentimes outperform their respective base models.
Constructing instruction datasets often involves using templates to transform text--label pairs into instruction--output pairs \citep{bigscience/xP3,longpre_flan_2023}. Our approach adopts this method for constructing IC and SF datasets for fine-tuning task-specific adapters with PEFT.

\vspace{2pt}
Despite prior work on IT, its impact on NLU tasks like intent classification and slot filling has been under-explored. Our work aligns with LINGUIST \cite{rosenbaum_linguist_2022}, which focuses on generating annotated data for IC and SF labels through instruction tuning. The generated data, however, is used to fine-tune a BERT-based model for joint IC+SF \citep{chen_bert_2019}, while we fine-tune Instruct-LLMs for IC- and SF-prompting.

%% file: methodology.tex
\paragraph{Problem Statement.} 
\label{sec:problem_statement}
We consider two NLU tasks where 
a single user utterance \texttt{\bf x}, with tokens $x_1, x_2, ..., x_n$, yields an output structure \texttt{\bf y}.
For example, given \texttt{\bf x}$=$\emph{``Find me an Italian restaurant with a parking lot''}, 
\texttt{\bf y} in intent classification (IC) is an intent label $l$ (e.g., \emph{find restaurant}) representing the user's intent in \texttt{\bf x}.
For slot filling (SF), \texttt{\bf y} comprises slot type--value pairs $\{(t_i,v_i)\}^{m}_{i=1}$, with $t_i$ as the slot type (e.g., \emph{cuisine}) and $v_i$ as the corresponding slot value (e.g., \emph{`Italian'}) extracted from \texttt{\bf x}.
In contrast to traditional slot filling where $v_i$ is always a span of \texttt{\bf x}, 
we also consider slots where
$v_i$ can be inferred from \texttt{\bf x}, e.g., (\emph{parking available, `yes'}) from \emph{``...with a parking lot''}. This scenario is present in dialog state tracking (DST) benchmark datasets, reflecting a more realistic downstream application.

\paragraph{ILLUMINER,} our approach for IC and SF using instruction-tuned LLMs, is illustrated in Figure~\ref{fig:IC-SF-prompting}. The input utterance \texttt{\bf x} is 
transformed into a task-specific prompt for either IC or SF.
This prompt is then provided to an Instruct-LLM, possibly enhanced by a task-specific PEFT adapter. Conditioned on the prompt, the Instruct-LLM generates \texttt{\bf y} specific to the task.

\vspace{2pt}
A prompt typically consists of an {\bf \emph{instruction}} distinguishing IC from SF prompts, and an {\bf \emph{input}} containing \texttt{\bf x}. For the in-context learning approach, the prompt also contains {\bf \emph{few-shot examples}} between the \emph{instruction} and the \emph{input}, in which each example utterance \texttt{\bf x$'$} follows the same template as \texttt{\bf x}, but is accompanied by the expected \texttt{\bf y$'$}. 
Note that providing few-shot examples to prompt an Instruct-LLM already enhanced with a PEFT adapter does not necessarily enhance performance and leads to longer inference time due to extended prompts; thus, they are never utilized in conjunction.

\paragraph{Prompt Engineering for IC.} 
We include the list of possible intent labels $L$ in the \emph{instruction}, derived from the ground-truth intents in the evaluation set of considered datasets. Instead of the original intent labels as annotated in the dataset, we employ handcrafted intent descriptions as labels, e.g., \emph{`turn light on'} (\emph{iot\_hue\_lighton}), \emph{`express liking music'} (\emph{music\_likeness}),
as they enhance label semantics and improve Instruct-LLMs' comprehension.
We explore four prompt template variations for IC (Table~\ref{tab:IC-prompts}), with $L$ listed in a single line per label.

\begin{table}[t]
    \centering
    \small
    \begin{adjustbox}{width=0.48\textwidth}
    \begin{tabular}{@{}lll@{}}
        \toprule
        $P_1$ & instruction & Given the possible intents: \{$L$\} \\
        \cmidrule{2-3}
         & input & What is the user's intent in '\texttt{\bf x}'? Intent: \\ 
        \toprule
        $P_2$ & instruction & Given the following options: \{$L$\} \\
        \cmidrule{2-3}
         & input & What did the user want when the user said, '\texttt{\bf x}'? Answer:\\
        \toprule
        $P_3$ & instruction & Classify the USER's utterances into one of the following\\
         & & intent options: \{$L$\} \\
        \cmidrule{2-3}
         & input & USER: '\texttt{\bf x}' Intent: \\
        \toprule
        $P_4$ & instruction & Given a USER's utterance, choose one of the following\\
         & & intents: \{$L$\} \\
        \cmidrule{2-3}
         & input & USER: '\texttt{\bf x}' Intent: \\
        \bottomrule
    \end{tabular}
    \end{adjustbox}
    \caption{Prompt template variations for IC.}
\label{tab:IC-prompts}
\end{table}

\paragraph{Prompt Engineering for SF.}
In the \emph{instruction} for SF, we expose the list of candidate slots $S$ in the form of \{$t_i$: $d_i$\}$^{m}_{i=1}$ where $t_i$ is a slot type (e.g., \emph{cuisine}) and $d_i$ is its corresponding description (e.g., \emph{`type of cuisine'}). Candidate slots $S$ are those relevant for the user's intent in a given utterance, e.g., \emph{cuisine} and \emph{price-range} for the \emph{find restaurant} intent. We derived relevant slot types based on intent--slot type co-occurrences (at least once) in the training data. When constructing \texttt{\bf y$'$} for the \emph{few-shot examples}, we insert \emph{null} for relevant slots not present in the ground truth slots. For example, with \texttt{\bf x$'$}$=$\emph{``I'd like to find a restaurant that serves Chinese food!''}, \texttt{\bf y$'$}$=$\{(\emph{cuisine}, \emph{`Chinese'}), (\emph{price-range}, \emph{null})\}.

In datasets like SNIPS and MASSIVE, annotations include general slot types like \emph{time} and \emph{city}. We designate general slots $S_{G}$ as slot types co-occurring with more than three intent labels. We incorporate $S_{G}$ alongside relevant slots $S$ as part of the \emph{instruction}.
Regarding prompt templates, we explore one variation illustrated in Figure~\ref{fig:IC-SF-prompting}, following several iterations of prompt design in a preliminary study.

%% file: experiments.tex
\section{Experimental Setup}
\label{sec:experimental-setup}



\begin{table}[t]
    \centering
    \small
    \begin{adjustbox}{width=0.48\textwidth}
    \begin{tabular}{@{}lccrrrr@{}}
        \toprule
        \multirow{3}{*}{\textbf{Dataset}} & \multirow{3}{*}{\textbf{\# Intents}} & \multirow{3}{*}{\textbf{\# Slots}} & \multicolumn{4}{c}{\textbf{Avg. prompt length}} \\
         &  &  & \multicolumn{2}{c}{\textbf{zero-shot}} & \multicolumn{2}{c}{\textbf{few-shot}} \\ 
         \cmidrule(lr){4-5} \cmidrule(lr{0em}){6-7}
         &  &  & \multicolumn{1}{c}{IC} & \multicolumn{1}{c}{SF} & \multicolumn{1}{c}{IC} & \multicolumn{1}{c}{SF} \\
        \toprule
        SNIPS & 7 & 45 & 75.4 & 115.2 & 294.4 & 493.8 \\
        MASSIVE & 60 & 55 & 336.7 & 160.5 & 603.6 & 551.9 \\
        MultiWoz & 11 & 24 & 83.9 & 90.8 & 450.9 & 303.5 \\
        \bottomrule
    \end{tabular}
    \end{adjustbox}
    \caption{Datasets for IC and SF experiments.}
\label{tab:dataset}
\end{table}

\paragraph{Dataset.}
We consider \textbf{(i)} \emph{SNIPS} \citeplanguageresource{coucke2018snips}, \textbf{(ii)} \emph{MASSIVE} \citeplanguageresource{fitzgerald2022massive} (English split) and \textbf{(iii)} \emph{MultiWoz 2.2} \citeplanguageresource{budzianowski2018multiwoz} as our benchmark datasets (see Table~\ref{tab:dataset}) since they encompass both IC and SF objectives and are widely used in the community. For MultiWoz, we consider only the first turn of each conversation in the test set for evaluation, as the first turn typically conveys a clear intent and precise slots in a single utterance, while subsequent turns may necessitate dialogue history for context.

\paragraph{Evaluation Metric.}
Following the baselines, we evaluate the performance of our proposed approach using the standard automatic evaluation metrics of \emph{accuracy} for IC and \emph{micro F1-score} for SF. 
We define \emph{hallucinations} for IC and SF as the ratio of false positives that cannot be found in candidate intent/slot labels and user utterances. 

\paragraph{Models.} 
We explore various Instruct-LLMs:

\begin{itemize}[leftmargin=*,topsep=0pt,itemsep=0pt,before=\vspace{0pt},after=\vspace{0pt}]
    \item \emph{Falcon-7B-Instruct} (\href{https://huggingface.co/tiiuae/falcon-7b-instruct}{tiiuae/falcon-7b-instruct}), Falcon-7B \citep{falcon40b} fine-tuned on a mixture of chat/instruct datasets \citeplanguageresource{penedo_refinedweb_2023,xu_baize_2023}.
    \item \emph{BLOOMZ} (\href{https://huggingface.co/bigscience/bloomz-7b1}{bigscience/bloomz-7b1}), fine-tuned BLOOM \citep{workshop_bloom_2022} on xP3 \citeplanguageresource{bigscience/xP3}, a collection of human-instruction datasets in 46 languages.
    \item \emph{FLAN-T5} (\href{https://huggingface.co/google/flan-t5-xxl}{google/flan-t5-xxl}), fine-tuned T5 (11B) on the Flan Collection \citeplanguageresource{longpre_flan_2023}, 
    $<$instruction, output$>$ pairs constructed from 62 datasets of 12 NLP tasks. 
    \item \emph{Vicuna} (\href{https://huggingface.co/lmsys/vicuna-13b-v1.5}{lmsys/vicuna-13b-v1.5}), from fine-tuning LLaMA 2 (13B, \citealt{touvron_llama_2023-1}) on 70K user-shared conversations collected from a website. 
    \item \emph{WizardLM} (\href{https://huggingface.co/WizardLM/WizardLM-13B-V1.1}{WizardLM/WizardLM-13B-V1.1}), fine-tuned LlaMA (13B, \citealt{touvron_llama_2023}) on the Evol-Instruct dataset \citeplanguageresource{xu2023wizardlm}.
\end{itemize}

We chose medium-sized LLMs (7B--13B) to compare various Instruct-LLMs of similar size but with different architectures and fine-tuned on distinct datasets. 
To restrict the generation, we set 10 as the maximum new tokens for IC and 100 for SF. 

    \paragraph{Zero-shot vs Few-shot.}
    In the few-shot setting, 
    prompts contain $k$ examples of
    user utterances and desired outputs (i.e., intent labels or slot type--value pairs), in contrast to the zero-shot setting.
    For intent classification, we randomly select one example per intent label from a small training set ($k$ examples per label, where $k=10$), forming the few-shot set $F$. Due to the limited context size of LLMs, we randomly sample 10 examples from $F$ when necessary.
    For slot filling, we randomly sample utterances from the training data until we fulfill the requirement of one example per slot type, yielding the few-shot set $F$, which we also constrain to a size of 10. Table~\ref{tab:dataset} details the average prompt length for IC and SF in zero- vs few-shot settings.
    \paragraph{Parameter-efficient Fine-Tuning.}
    We explore various PEFT techniques including prefix-tuning, prompt-tuning, LoRA and \iathree, implemented by Hugging Face\footnote{{\url{https://github.com/huggingface/peft}}}.
    We train the PEFT adapters separately for IC and SF on each dataset using a small training set ($k$ examples per label, where $k=10$). 
    Adapters for IC are fine-tuned by varying the prompt templates as listed in Table~\ref{tab:IC-prompts}.
    In both prefix and prompt tuning, we learn 20 virtual tokens with a learning rate of $1\mathrm{e}{-2}$. In the case of LoRA, we set the hyper-parameters as $r=16$, $\mathrm{lora\_alpha}=32$ and $\mathrm{lora\_dropout}=0.1$.
    We optimize the learning rate for Lora and \iathree by selecting from $\{5\mathrm{e}{-4}, 1\mathrm{e}{-3}, 5\mathrm{e}{-3}\}$, and the number of epochs from $\{5, 10, 20\}$.
    We employ $AdamW$ as the optimizer with its default hyper-parameters.
    


\paragraph{Baselines.}
\label{sec:baselines}
We consider the following baselines for IC and SF tasks: 

\begin{itemize}[leftmargin=*,topsep=0pt,itemsep=0pt]
    \item \emph{JointBERT}, BERT-based models (110M) for joint IC+SF \citep{chen_bert_2019}.\footnote{\url{https://github.com/monologg/JointBERT}} Using default hyperparameters (batch size of 32, learning rate of $5\mathrm{e}{-5}$), we train models for 20 epochs in experiments with full training data. In experiments with a small training set ($k=10$ per label), we reduced the batch size to 8 and train models for 50 epochs.
    \item OpenAI \emph{GPT3.5} (text-davinci-003), 175B GPT3 \citep{brown_language_2020} that has been trained on a larger dataset, enhancing its capability on understanding natural language instructions.
    \item \emph{LINGUIST} \citep{rosenbaum_linguist_2022}, which leverages instruction-tuned LLMs (AlexaTM-5B) to generate annotated data for IC and SF, given few-shot examples as seed set. The generated data is used to fine-tune a BERT-style model for joint IC+ST \citep{chen_bert_2019}.
\end{itemize}

\begin{table*}[!t] 
\centering 
\small
\begin{adjustbox}{width=1.0\textwidth}
\begin{tabular}{@{}llccccccccc@{}}
\toprule
\multirow{2}{*}{\bf Instruct-LLM} & \multirow{2}{*}{\bf Size} & \multicolumn{3}{c}{\bf SNIPS} & \multicolumn{3}{c}{\bf MASSIVE} & \multicolumn{3}{c}{\bf MultiWoz} \\
\cmidrule(lr){3-5} \cmidrule(lr){6-8} \cmidrule(lr{0em}){9-11}
 & & zero-shot & few-shot & LoRA & zero-shot & few-shot & LoRA & zero-shot & few-shot & LoRA \\
 \midrule
falcon-7b-instruct & 7B & .301 $\pm$.15 & .570 $\pm$.03 & .779 $\pm$.06 & .103 $\pm$.05 & .360 $\pm$.03 & .546 $\pm$.00 & .558 $\pm$.38 & .748 $\pm$.01 & .941 $\pm$.02 \\
bloomz-7b1 & 7B & .795 $\pm$.08 & .686 $\pm$.10 & .930 $\pm$.01 & .265 $\pm$.06 & .435 $\pm$.02 & .657 $\pm$.01 & .899 $\pm$.02 & .894 $\pm$.06 & .941 $\pm$.01 \\
flan-t5-xxl & 11B & {\bf .937 $\pm$.01} & {\bf .940 $\pm$.00} & {\bf .962 $\pm$.00} & {\bf .726 $\pm$.01} & {\bf .741 $\pm$.01} & {\bf .825 $\pm$.01} & {\bf .973 $\pm$.00} & {\bf .982 $\pm$.00} & {\bf .979 $\pm$.00} \\
vicuna-13b-v1.5 & 13B & .574 $\pm$.30 & .920 $\pm$.01 & .950 $\pm$.01 & .333 $\pm$.23 & .688 $\pm$.01 & .759 $\pm$.01 & .425 $\pm$.17 & .977 $\pm$.00 & .972 $\pm$.01 \\
WizardLM-13B-V1.1 & 13B & .720 $\pm$.25 & .674 $\pm$.20 & .921 $\pm$.01 & .355 $\pm$.11 & .678 $\pm$.01 & .731 $\pm$.01 & .962 $\pm$.02 & .933 $\pm$.02 & .956 $\pm$.01 \\
\bottomrule
\end{tabular}
\end{adjustbox}
\caption{Intent accuracy in zero-shot, few-shot and LoRA settings, across different datasets and Instruct-LLMs. Numbers following $\pm$ indicate standard deviation across different prompts.}
\label{tab:IC}
\end{table*}

\begin{table*}[!t] 
\centering 
\small
\begin{adjustbox}{width=1.0\textwidth}
\begin{tabular}{@{}llccccccccc@{}}
\toprule
\multirow{2}{*}{\bf Instruct-LLM} & \multirow{2}{*}{\bf Size} & \multicolumn{3}{c}{\bf SNIPS} & \multicolumn{3}{c}{\bf MASSIVE} & \multicolumn{3}{c}{\bf MultiWoz} \\
\cmidrule(lr){3-5} \cmidrule(lr){6-8} \cmidrule(lr{0em}){9-11}
 & & zero-shot & few-shot & LoRA & zero-shot & few-shot & LoRA & zero-shot & few-shot & LoRA \\
\midrule
\shaderow \multicolumn{11}{@{}l@{}}{ILLUMINER: Single-prompt IE}\\
falcon-7b-instruct & 7B & .136 (.70) & .543 (.18) & .835 (.01) & .042 (.87) & .421 (.29) & .585 (.02) & .319 (.66) & .640 (.24) & .928 (.02) \\
bloomz-7b1 & 7B & .177 (.66) & .541 (.19) & .876 (.00) & .043 (.81) & .349 (.30) & .640 (.01) & .278 (.70) & .527 (.20) & .943 (.01) \\
flan-t5-xxl & 11B & {\bf .310 (.35)} & .647 (.14) & {\bf .909 (.01)} & {\bf .125 (.37)} & .473 (.21) & {\bf .735 (.00)} & .462 (.46) & .753 (.18) & .945 (.02) \\
vicuna-13b-v1.5 & 13B & .222 (.43) & .554 (.08) & .908 (.01) & .103 (.59) & .369 (.14) & .724 (.03) & {\bf .500 (.36)} & {\bf .859 (.10)} & {\bf .957 (.01)} \\
WizardLM-13B-V1.1 & 13B & .298 (.53) & {\bf .685 (.12)} & .899 (.00) & .116 (.67) & {\bf .474 (.19)} & .710 (.01) & .428 (.53) & .830 (.10) & .951 (.02) \\
\midrule
\shaderow \multicolumn{11}{@{}l@{}}{Multi-prompt IE \citep{hou_inverse_2022}}\\
flan-t5-xxl & 11B & \uline{.221 (.54)} & .380 (.22) & \uline{.904 (01)} & \uline{.058 (.74)} & .195 (.45) & \uline{.658 (.00)} & \uline{.360 (.47)} & .547 (.31) & \uline{.933 (.02)} \\
vicuna-13b-v1.5 & 13B & .111 (.73) & \uline{.569 (.06)} & .755 (.01) & .021 (.90) & \uline{.252 (.29)} & .597 (.04) & .146 (.89) & \uline{.798} (.09) & .889 (.05) \\
WizardLM-13B-V1.1 & 13B & .127 (.69) & .531 (.13) & .703 (.01) & .020 (.93) & .202 (.42) & .509 (.01) & .255 (.76) & .717 (.17) & .855 (.08) \\
\bottomrule
\end{tabular}
\end{adjustbox}
\caption{Slot filling F1 in zero-shot, few-shot and LoRA settings, across different datasets and Instruct-LLMs. Numbers inside parentheses indicate the ratio of wrong predictions caused by hallucinations.}
\label{tab:SF}
\end{table*}

\section{Results and Analysis}

\paragraph{Intent Classification.} Table~\ref{tab:IC} summarizes the performance of various Instruct-LLMs across different settings for all datasets considered. FLAN-T5 (flan-t5-xxl) consistently outperforms other models, including slightly larger ones like Vicuna (vicuna-13b-v1.5) and WizardLM (WizardLM-13B-V1.1). We suggest that for intent classification, encoder-decoder models, such as FLAN-T5, excel in capturing utterance meaning, leading to superior sequence classification performance. 
LoRA fine-tuning outperforms few-shot learning in most cases, especially for 7B models and on MASSIVE where the set of intent labels is significantly larger.
Fine-tuning 7B models yields comparable results to larger models on SNIPS and MultiWoz, but the gap widens on the challenging MASSIVE. 
Examining standard deviation across different prompt templates (Table~\ref{tab:IC-prompts}), FLAN-T5 emerges as the most robust model, with a standard deviation $\le 0.01$. Notably, fine-tuning with LoRA also helps in reducing performance variance when using different prompts.

\paragraph{Slot Filling.} We evaluate Instruct-LLMs for slot filling, taking into account ground truth intent labels in the prompt construction and few-shot example generation. As reported in Table~\ref{tab:SF}, FLAN-T5, Vicuna and WizardLM exhibit competitive performance, with no clear winner. However, in the few-shot setting, Vicuna outperforms FLAN-T5 on MultiWoz whereas WizardLM on SNIPS and MASSIVE datasets, suggesting that causal decoder models excel when the generation capability is essential for producing structured outputs like slot type--value pairs. Fine-tuning with LoRA significantly improves performance by reducing false positives, i.e., the models learn how and when to fill slots with \emph{null} values. Furthermore, the few-shot setting and LoRA mitigate hallucinations (indicated by numbers inside parentheses) considerably, as they help controlling the models' behavior to only fill the slots with relevant information found in the input utterances. 

In a comparison with prior work \citep{hou_inverse_2022}, where FLAN-T5, Vicuna and WizardLM are prompted using the multi-prompt IE strategy, we find that the best results from this technique (Table~\ref{tab:SF}, underlined) are inferior to the best results from our prompting approach (Table~\ref{tab:SF}, bold) for slot filling across all settings and datasets.

\begin{table}[!t] 
\centering 
\small
\begin{adjustbox}{width=0.48\textwidth}
\begin{tabular}{@{}lc@{\hspace{4pt}}cc@{\hspace{4pt}}cc@{\hspace{4pt}}c@{}}
\toprule
\multirow{2}{*}{\bf Model} & \multicolumn{2}{c}{\bf SNIPS} & \multicolumn{2}{c}{\bf MASSIVE} & \multicolumn{2}{c}{\bf MultiWoz} \\
\cmidrule(lr){2-3} \cmidrule(lr){4-5} \cmidrule(lr@{0em}){6-7}
 & IC & SF & IC & SF & IC & SF \\
 \midrule
 \shaderow \multicolumn{7}{@{}l@{}}{$k=10$ per label} \\
 ILLUMINER ($\mbox{flan-t5-xxl}_{\scriptsize \mbox{~LoRA}}$) & \uline{.961} & \uline{.899} & \uline{.833} & \uline{.720} & .978 & \uline{.946} \\
 ILLUMINER ($\mbox{flan-t5-xxl}_{\scriptsize \mbox{~few-shot}}$) & .918 & .600 & .718 & .440 & .970 & .746 \\
 JointBERT \citep{chen_bert_2019} & .907 & .608 & .718 & .609 & .958 & .747 \\
 $\mbox{GPT3.5}_{\scriptsize \mbox{~zero-shot}}$ & .913 & .487 & .716 & .372 & \uline{.979} & .696 \\
 $\mbox{GPT3.5}_{\scriptsize \mbox{~few-shot}}$ & .931 & .633 & .757 & .398 & .973 & .831 \\
 \citet{rosenbaum_linguist_2022} $\dagger$ & .920 & .823 & - & - & - & - \\
\midrule
 \shaderow \multicolumn{7}{@{}l@{}}{Full training set} \\
 ILLUMINER ($\mbox{flan-t5-xxl}_{\scriptsize \mbox{~LoRA}}$) & .967 & .948 & .871 & {\bf .797} & .989 & {\bf .962} \\
 JointBERT \citep{chen_bert_2019} & {\bf .983} & {\bf .965} & {\bf .885} & {\bf .797} & {\bf .990} & .834 \\
\bottomrule
\end{tabular}
\end{adjustbox}
\caption{Comparison with baselines in terms of intent accuracy (IC) and slot filling F1 (SF). $\dagger$ denotes that numbers were taken directly from the paper.}
\label{tab:Joint-IC-SF}
\end{table}

\paragraph{Comparison with Baselines.} Here we consider the task of joint IC and SF, where the predicted intents are used for building the prompt for SF (determining candidate slots $S$), differing from previous SF experiments that used ground-truth intent labels. To address potential error propagation from IC to SF, we selected $\mbox{flan-t5-xxl}_{\scriptsize \mbox{~LoRA}}$ to instantiate ILLUMINER, given its superior performance in IC (Table~\ref{tab:IC}), coupled with the IC prompt template $P_1$. We present its performance against considered baselines (\S~\ref{sec:baselines}) in Table~\ref{tab:Joint-IC-SF}.

Small LMs (e.g., BERT, 110M) outperform larger models when fine-tuned on full training data, as indicated in bold in Table~\ref{tab:Joint-IC-SF}. However, with a small training set ($k=10$ per label, 0.5\%--5.2\% of the full training set), ILLUMINER ($\mbox{flan-t5-xxl}_{\scriptsize \mbox{~LoRA}}$) offers clear advantages over all baselines, delivering the best performance (underlined in Table~\ref{tab:Joint-IC-SF}), particularly on challenging datasets like MASSIVE (with 60 intent labels) and intricate tasks like slot filling. 
Medium-sized fine-tuned Instruct-LLMs (e.g., $\mbox{flan-t5-xxl}_{\scriptsize \mbox{~LoRA}}$, 11B) even outperform few-shot learning with much larger models (e.g., $\mbox{GPT3.5}_{\scriptsize \mbox{~few-shot}}$, 175B), highlighting the significance of exposing LLMs to more comprehensive examples, even if limited in size, achievable with parameter-efficient fine-tuning (PEFT). Provided with the same few-shot examples as $\mbox{GPT3.5}_{\scriptsize \mbox{~few-shot}}$, ILLUMINER ($\mbox{flan-t5-xxl}_{\scriptsize \mbox{~few-shot}}$) exhibits lower performance in most cases, although it remains comparable especially for intent classification.

LINGUIST \citep{rosenbaum_linguist_2022} performs similarly on SNIPS with its data augmentation method. Yet, approaches relying on \emph{sequence labelling/tagging} for SF (e.g., JointBERT, LINGUIST) have limited capabilities on MultiWoz, where 13.8\% of slot values are \emph{inferred} from input utterances, capping recall at 0.862. This underscores the superiority of our SF-prompting method with ILLUMINER.

\section{Ablation Studies}

To better study and analyze the effectiveness of task-adapted instruction tuning with PEFT, we conduct the following series of ablation experiments:

\input{images/ablation_instruct_non-instruct}
\paragraph{Instruct- vs Non-instruct LLMs.}



For this study, we examine FLAN-T5-large (780M), BLOOMZ (7B) and Falcon-Instruct (7B) as Instruct-LLMs, with T5-large, BLOOM and Falcon as their non-instruct counterparts.
We fine-tune and evaluate LoRA adapters for these models, and present the results in Figure~\ref{fig:ablation-instruct-vs-non-instruct}.
Our observations indicate that FLAN-T5 consistently exhibits superior performance over T5 in both tasks, showcasing its adept learning of task-specific instructions. BLOOMZ also demonstrates improved performance compared to its base counterpart in most cases, except when classifying intents within the MASSIVE dataset.
However, the Falcon models present a unique scenario where Falcon-Instruct, in contrast to other instruct models, does not consistently outperform the non-instruct base version. 
From Figure~\ref{fig:ablation-instruct-vs-non-instruct}, it is evident that the MASSIVE dataset is the most challenging to solve. 
This variation may stem from the complexity of the dataset, posing challenges for the instruct model to generalize effectively to intricate patterns and implicit dependencies. This points to the need for further investigations and improved hyper-parameter tuning of instruction-tuned models.

\input{images/ablation_model_size}
\paragraph{Varying Model Size.}
We assess the impact of varying the model size for ILLUMINER instantiated with $\mbox{FLAN-T5}_{\scriptsize \mbox{~LoRA}}$, and report the results in Figure~\ref{fig:ablation-model-size}. While there are notable gains with increased model size, especially for SF on all datasets and for IC on MASSIVE, smaller models perform nearly as well as the largest one for IC on SNIPS and MultiWoz. This indicates that larger models excel in tasks with a vast set of labels. However, we observe a diminishing trend in performance gains after 3B, suggesting that leveraging models larger than 11B in the PEFT setting likely offers no advantages.

\input{images/ablation_num_examples_per_label}
\paragraph{Varying Number of Examples per Label.}
In Figure~\ref{fig:ablation-num-examples-per-label}, increasing the number of examples ($k$) per label for ILLUMINER with $\mbox{FLAN-T5-xxl}_{\scriptsize \mbox{~LoRA}}$ shows minimal performance gains, except for SF on SNIPS and MASSIVE, where using all training instances leads to 5.2 and 7.7 percentage points improvement. This demonstrates that Instruct-LLMs in the PEFT setting are able to generalize effectively even with extremely limited fine-tuning data.

\input{images/ablation_peft_techniques}
\input{images/ablation_peft_generalization}

\paragraph{Different PEFT techniques.}

As illustrated in Figure \ref{fig:ablation-peft-techniques}, FLAN-T5-xxl was fine-tuned with different PEFT techniques to compare the trends. All the techniques were trained with parameters amounting to less than 0.1\% of the model parameters. For the relatively easier IC task, the four techniques offer comparable performance across datasets. However, LoRA stands out as the most suitable for SF across all datasets. 
Notably, prefix- and prompt-tuned models exhibit poor SF performance in SNIPS and MASSIVE but perform similarly to LoRA and \iathree in MultiWoZ.

\paragraph{Generalization across datasets.}
We extended our evaluation across datasets for the models trained with diverse PEFT techniques to learn their cross-dataset generalization capabilities. For instance, LoRA adapters for FLAN-T5-xxl were trained individually on SNIPS for IC and SF, and evaluated on MASSIVE and MultiWoZ. Similar evaluations were conducted for all considered PEFT techniques and across all combinations of train/eval datasets. In Figure \ref{fig:ablation-peft-generalization}, we report generalization trends with average Accuracy for IC and average F1 for SF in the cross-dataset evaluation. It is evident that LoRA and \iathree offer impressive generalization over cross-dataset evaluation in both tasks. While prompt tuning shows comparable IC generalization, it exhibits significantly lower performance in SF generalization.

\input{images/ablation-exposure-labels}
\input{images/ablation-multilinguality}
\input{tables/discussion}

\paragraph{Exposure of labels in instructions.}

To study the impact of the inclusion of labels ($L$ and $S$ for IC and SF, respectively, as defined in \S~\ref{sec:problem_statement}) in instruction tuning, we conducted additional fine-tuning and evaluation of the FLAN-T5-xxl model without exposing these labels in the instruction. 
As shown in Table \ref{tab:exposure_labels}, models fine-tuned with instructions containing $L$ and $S$ (\textsc{train}$+$) outperform those without these labels during fine-tuning (\textsc{train}$-$). This difference is especially pronounced in SF, 
where it exceeds 62\% on average.
Interestingly, models trained with $L$ and $S$ perform relatively poorly when these labels are excluded during evaluation (\textsc{test}$-$). 
We conclude that fine-tuning models for IC and SF with labels in the instruction 
enables the generation of output labels from candidates $L$ and $S$, offering improved generalization across domains and datasets compared to learning solely from model weights and data distributions.

\paragraph{Multilinguality.}
To investigate the applicability of our ILLUMINER framework for languages beyond English, we performed IC and SF experiments across five language splits within the MASSIVE dataset: English (en), German (de), French (fr), Italian (it), and Spanish (es).
In Table \ref{tab:multilinguality}, we report the performance of ILLUMINER instantiated with the following LoRA fine-tuned models:
\vspace{3pt}
\begin{itemize}[leftmargin=*,topsep=0pt,itemsep=-2pt]
\item \emph{FLAN-T5-xxl}, trained with mostly English texts.
\item \emph{mT5-xxl} \cite{xue_mt5_2021}, a multilingual variant of T5 covering 101 languages.
\item \emph{mT0-xxl(-mt)}, mT5 fine-tuned on a cross-lingual instruction dataset, xP3 \citeplanguageresource{bigscience/xP3}. The \emph{-mt} variant is recommended for prompting in non-English.
\end{itemize}
\vspace{3pt}
Multilingual LLMs generally exhibit lower performance on the English split compared to FLAN-T5. However, apart from IC on the German split, we observe the advantages of utilizing multilingual LLMs ($\mbox{mt5-xxl}_{\scriptsize \mbox{~LoRA}}$ and $\mbox{mt0-xxl}_{\scriptsize \mbox{~LoRA}}$) for non-English input utterances, even when the task instructions and label descriptions are still in English.
Instruction-tuned mT5, referred to as mT0, demonstrates superior performance across all considered languages, validating our previous observation that applying PEFT on Instruct-LLMs yields greater benefits.
When employing $\mbox{mt0-xxl-mt}_{\scriptsize \mbox{~LoRA}}$, we translated both task instructions and label descriptions into the respective languages of the input utterances, during both fine-tuning and inference stages. While we observe performance increase in SF when the prompts were translated, the same improvement was not always evident for IC. We conjecture that this discrepancy arises from translations often yielding longer and more ambiguous label descriptions, particularly noticeable for French.

\section{Discussion}

Based on the experimental outcomes, we record a few observations as shown in Table \ref{tab:discussion}, discussing the shortcomings and advantages of few-shot learning and instruction tuning for IC and SF.

    \paragraph{Ambiguous User Utterances.} In TOD systems, models often deal with ambiguous user utterances as input, facing challenges in accurately identifying potential intents.
    Example 1 illustrates such an utterance, where the annotated intent is related to `pink' as the smart lighting's color, while ILLUMINER misunderstands it as a singer and GPT3.5 falls back to the out-of-scope intent (\emph{be quirky}).
    Users convey intents in numerous ways,
    posing difficulties for models to generalize across variations.

    \paragraph{Entity Disambiguation.} 
    In many cases including Example 1 and 2, words and phrases may refer to different entity types, requiring models to disambiguate them in the given context. However, given only few samples for either few-shot learning or fine-tuning, it is often hard for LLMs to understand patterns or guidelines employed by human annotators on deciding slot labels (e.g., `australian' \textit{time-zone} against the place `australia').

    \paragraph{Missing Context.} 
    Single-turn IC and SF is highly challenging due to limited context as compared to a multi-turn setting with previous turns in the conversation as context.
    In Example 3, 
    context absence hinders the models to predict the expected intent. 

    \paragraph{Highly Correlated Labels.} 
    Highly correlated labels where the distinctions are often subtle and context-dependent, such as \textit{entity-name} and \textit{artist}, making it challenging to precisely predict intents and slots given user utterances.
    This also points towards data inconsistencies 
    and label noise in large datasets.
    Nevertheless, ILLUMINER correctly identified \emph{`lindsey cardinale'} in Example 4 as \emph{artist},
    supporting the hypothesis that fine-tuning may resolve such problems for most examples, if not entirely.

     \paragraph{Hallucinations.} LLMs are prone to hallucinations, as evidenced in our use case where they generate intents and slots absent in candidate labels or user utterances.
     Example 5 and 6 depict such a \emph{factual mirrage}~\cite{rawte_troubling_2023} for IC and SF, respectively, where ILLUMINER generated the \emph{turn an alarm on} intent not present in the candidate labels, and GPT3.5 generated the \emph{player-setting}: \emph{`repeat'} slot when \emph{`repeat'} is never mentioned.
     Approximately 2.94\% of false positives for IC and 3.76\% for SF, with ILLUMINER, fall into this error category.

    \vspace{5pt} 
    
For future research, we plan to extend our study to multi-turn settings to tackle context deficiency. Techniques like semantic-driven label mapping, confidence scoring for prediction reliability assessment, and requesting clarification could mitigate hallucination risks.

\section{Conclusion}
We introduced ILLUMINER for intent classification (IC) and slot filling (SF) with Instruct-LLMs. Our LoRA fine-tuned models surpass GPT3.5 in zero- and few-shot settings, as well as the state-of-the-art joint IC+SF approach. Notably, we achieve impressive results using less than 6\% of the training data across benchmarks like SNIPS, MASSIVE and MultiWoZ. These findings have direct practical applications in task-oriented dialogue systems, enabling enhanced performance with reduced computational power and data annotation efforts.

\section*{Acknowledgement}
This research was funded by the German Federal Ministry for Economic Affairs and Climate Action (BMWK) through the project OpenGPT-X (project no. 68GX21007D).


%% file: images/ablation_instruct_non-instruct.tex
\pgfplotstableread{
model           SNIPS-IC    MASSIVE-IC  MultiWoz-IC SNIPS-SF    MASSIVE-SF  MultiWoz-SF
t5              0.911       0.776       0.978       0.804       0.668       0.942     
flan-t5         0.94        0.786       0.986       0.876       0.69        0.944
}\tfivetable

\pgfplotstableread{
model           SNIPS-IC    MASSIVE-IC  MultiWoz-IC SNIPS-SF    MASSIVE-SF  MultiWoz-SF
bloom           0.928       0.732       0.957       0.809       0.703       0.898
bloomz          0.927       0.712       0.971       0.887       0.714       0.943
}\bloomtable

\pgfplotstableread{
model           SNIPS-IC    MASSIVE-IC  MultiWoz-IC SNIPS-SF    MASSIVE-SF  MultiWoz-SF
falcon          0.759       0.677       0.93        0.758       0.704       0.909
falcon-instruct 0.767       0.595       0.946       0.834       0.652       0.928
}\falcontable

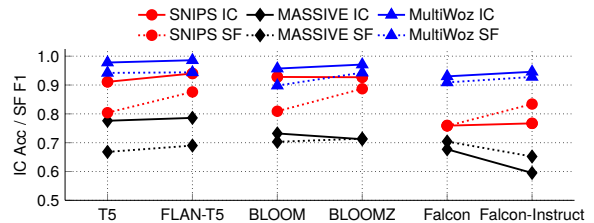
\begin{figure}[t]
    \flushleft
    \pgfplotsset{
       width=0.8\textwidth,
       height=0.2\textheight,
       grid=major,
       major grid style={dotted,draw=black!100},
       symbolic x coords={t5, flan-t5, bloom, bloomz, falcon, falcon-instruct},
       enlarge y limits={upper,value=0.05},
       legend style={
          draw=none,
          at={(0.5,1.0)},
          legend columns=3,
          legend cell align=left,
          anchor=south
          },
       }
    \begin{adjustbox}{width=0.48\textwidth}
    \begin{tikzpicture}
    \begin{axis}[
       axis y line*=left,
       axis x line*=bottom,
       xticklabels={T5, FLAN-T5, BLOOM, BLOOMZ, Falcon, Falcon-Instruct},
       yticklabel pos=left,
       ylabel near ticks,
       ymin=0.5, ymax=1.0,
       ytick={0.4, 0.5, 0.6, 0.7, 0.8, 0.9, 1.0},
       yticklabels={0.4, 0.5, 0.6, 0.7, 0.8, 0.9, 1.0},
       ylabel={IC Acc / SF F1},
       xtick=data,
       xticklabel style={
          inner sep=5pt,
          }
       ]

       \addplot[very thick,draw=red!100,mark=*,only marks,forget plot,mark options={scale=1.5,solid,fill=red}] plot coordinates{
          (t5,0.911) (flan-t5,0.94) (bloom,0.928) (bloomz,0.927) (falcon,0.76) (falcon-instruct,0.767)
          };
       
       \addplot[very thick,draw=red!100,mark=*,mark options={scale=1.5,solid,fill=red}] 
       table [x=model, y=SNIPS-IC] {\tfivetable};
       \addlegendentry{SNIPS IC}
       
       \addplot[very thick,draw=black!100,mark=diamond*,mark options={scale=1.8,solid,fill=black}] 
       table [x=model, y=MASSIVE-IC] {\tfivetable};
       \addlegendentry{MASSIVE IC}
       
       \addplot[very thick,draw=blue!100,mark=triangle*,mark options={scale=1.8,solid,fill=blue}] 
       table [x=model, y=MultiWoz-IC] {\tfivetable};
       \addlegendentry{MultiWoz IC}
       
       \addplot[very thick,draw=red!100,dotted,mark=*,mark options={scale=1.5,solid,fill=red}] 
       table [x=model, y=SNIPS-SF] {\tfivetable};
       \addlegendentry{SNIPS SF}
       
       \addplot[very thick,draw=black!100,dotted,mark=diamond*,mark options={scale=1.8,solid,fill=black}] 
       table [x=model, y=MASSIVE-SF] {\tfivetable};
       \addlegendentry{MASSIVE SF}
       
       \addplot[very thick,draw=blue!100,dotted,mark=triangle*,mark options={scale=1.8,solid,fill=blue}] 
       table [x=model, y=MultiWoz-SF] {\tfivetable};
       \addlegendentry{MultiWoz SF}

       \addplot[very thick,draw=red!100,mark=*,mark options={scale=1.5,solid,fill=red}] 
       table [x=model, y=SNIPS-IC] {\bloomtable};
       \addplot[very thick,draw=red!100,mark=*,mark options={scale=1.5,solid,fill=red}] 
       table [x=model, y=SNIPS-IC] {\falcontable};

       \addplot[very thick,draw=black!100,mark=diamond*,mark options={scale=1.8,solid,fill=black}] 
       table [x=model, y=MASSIVE-IC] {\bloomtable};
       \addplot[very thick,draw=black!100,mark=diamond*,mark options={scale=1.8,solid,fill=black}] 
       table [x=model, y=MASSIVE-IC] {\falcontable};

       \addplot[very thick,draw=blue!100,mark=triangle*,mark options={scale=1.8,solid,fill=blue}] 
       table [x=model, y=MultiWoz-IC] {\bloomtable};
       \addplot[very thick,draw=blue!100,mark=triangle*,mark options={scale=1.8,solid,fill=blue}] 
       table [x=model, y=MultiWoz-IC] {\falcontable};

       \addplot[very thick,draw=red!100,dotted,mark=*,mark options={scale=1.5,solid,fill=red}] 
       table [x=model, y=SNIPS-SF] {\bloomtable};
       \addplot[very thick,draw=red!100,dotted,mark=*,mark options={scale=1.5,solid,fill=red}] 
       table [x=model, y=SNIPS-SF] {\falcontable};

       \addplot[very thick,draw=black!100,dotted,mark=diamond*,mark options={scale=1.8,solid,fill=black}] 
       table [x=model, y=MASSIVE-SF] {\bloomtable};
       \addplot[very thick,draw=black!100,dotted,mark=diamond*,mark options={scale=1.8,solid,fill=black}] 
       table [x=model, y=MASSIVE-SF] {\falcontable};

       \addplot[very thick,draw=blue!100,dotted,mark=triangle*,mark options={scale=1.8,solid,fill=blue}] 
       table [x=model, y=MultiWoz-SF] {\bloomtable};
       \addplot[very thick,draw=blue!100,dotted,mark=triangle*,mark options={scale=1.8,solid,fill=blue}] 
       table [x=model, y=MultiWoz-SF] {\falcontable};
       
       \end{axis}
       
    \end{tikzpicture}
    \end{adjustbox}
    \caption{Instruct-LLMs vs their corresponding base models (non-instruct).}
    \label{fig:ablation-instruct-vs-non-instruct}
\end{figure}

%% file: images/ablation_model_size.tex
\pgfplotstableread{
size    SNIPS-IC    SNIPS-SF    MASSIVE-IC  MASSIVE-SF  MultiWoz-IC MultiWoz-SF
small   0.874       0.508       0.525       0.419       0.966       0.741       
base    0.834       0.69        0.702       0.544       0.958       0.912
large   0.94        0.848       0.786       0.669       0.966       0.93
xl      0.963       0.888       0.807       0.71        0.981       0.95
xxl     0.957       0.896       0.837       0.72        0.986       0.952
}\evaltable

\pgfplotstableread{
size    SNIPS-IC    SNIPS-SF    MASSIVE-IC  MASSIVE-SF  MultiWoz-IC MultiWoz-SF
small   35          54          93          83          37          53   
base    35          84          162         124         41          80
large   50          183         424         320         63          169
xl      82          454         1128        887         100         409
xxl     360         1309        5068        4808        461         1103
}\trainingtimetable

\begin{figure}[t]
    \centering
    \pgfplotsset{
       width=0.8\textwidth,
       height=0.2\textheight,
       grid=major,
       major grid style={dotted,draw=black!100},
       symbolic x coords={small, base, large, xl, xxl},
       enlarge y limits={upper,value=0.05},
       legend style={
          draw=none,
          at={(0.5,1.0)},
          legend columns=3,
          legend cell align=left,
          anchor=south
          },
       }
    \begin{adjustbox}{width=0.48\textwidth}
    \begin{tikzpicture}
    \begin{axis}[
       xtick pos=left,
       axis y line*=right,
       ybar,
       axis x line*=bottom,
       yticklabel pos=right,
       ylabel near ticks,
       ymajorgrids,
       bar width=0.5cm,
       ymin=0, ymax=6000,
       ytick={0, 1000, 2000, 3000, 4000, 5000, 6000},
       yticklabels={0, 1k, 2k, 3k, 4k, 5k, 6k},
       ylabel style={align=center},
       ylabel={Adapter-tuning time (seconds)},
       xtick=data,
       xticklabels={},
       ]
       \addplot[ybar,draw=none,ybar legend,fill=red!30] 
       table [x=size, y=SNIPS-IC] {\trainingtimetable};
       
       \addplot[ybar,draw=none,ybar legend,fill=black!30] 
       table [x=size, y=MASSIVE-IC] {\trainingtimetable};
       
       \addplot[ybar,draw=none,ybar legend,fill=blue!30] 
       table [x=size, y=MultiWoz-IC] {\trainingtimetable};
    \end{axis}

    \begin{axis}[
       ybar,
       bar width=0.5cm,
       ymin=0, ymax=6000,
       xtick=data,
       yticklabels={,,},
       xticklabels={},
       ]
       \addplot[ybar,draw=red!30,ybar legend,pattern=north east lines,pattern color=red!80] 
       table [x=size, y=SNIPS-SF] {\trainingtimetable};
       
       \addplot[ybar,draw=black!30,ybar legend,pattern=north east lines,pattern color=black!80] 
       table [x=size, y=MASSIVE-SF] {\trainingtimetable};
       
       \addplot[ybar,draw=blue!30,ybar legend,pattern=north east lines,pattern color=blue!80] 
       table [x=size, y=MultiWoz-SF] {\trainingtimetable};
    \end{axis}
    
    \begin{axis}[
       axis y line*=left,
       axis x line*=bottom,
       xticklabels={small (80M), base (250M), large (780M), xl (3B), xxl (11B)},
       yticklabel pos=left,
       ylabel near ticks,
       ymin=0.4, ymax=1.0,
       ytick={0.4, 0.5, 0.6, 0.7, 0.8, 0.9, 1.0},
       yticklabels={0.4, 0.5, 0.6, 0.7, 0.8, 0.9, 1.0},
       ylabel={IC Acc. / SF F1},
       xtick=data,
       xticklabel style={
          inner sep=5pt,
          }
       ]

       \addlegendimage{draw=red!30,/pgfplots/refstyle=SNIPS-IC}
       \addlegendentry{SNIPS}
       \addlegendimage{draw=black!30,/pgfplots/refstyle=MASSIVE-IC}
       \addlegendentry{MASSIVE}
       \addlegendimage{draw=blue!30,/pgfplots/refstyle=MultiWoz-IC}
       \addlegendentry{MultiWoz}
       
       \addplot[very thick,draw=red!100,mark=*,mark options={scale=1.5,solid,fill=red}] 
       table [x=size, y=SNIPS-IC] {\evaltable};
       \addlegendentry{SNIPS IC}
       
       \addplot[very thick,draw=black!100,mark=diamond*,mark options={scale=1.8,solid,fill=black}] 
       table [x=size, y=MASSIVE-IC] {\evaltable};
       \addlegendentry{MASSIVE IC}
       
       \addplot[very thick,draw=blue!100,mark=triangle*,mark options={scale=1.8,solid,fill=blue}] 
       table [x=size, y=MultiWoz-IC] {\evaltable};
       \addlegendentry{MultiWoz IC}
       
       \addplot[very thick,draw=red!100,dotted,mark=*,mark options=
       {scale=1.5,solid,fill=red}] 
       table [x=size, y=SNIPS-SF] {\evaltable};
       \addlegendentry{SNIPS SF}
       
       \addplot[very thick,draw=black!100,dotted,mark=diamond*,mark options={scale=1.8,solid,fill=black}] 
       table [x=size, y=MASSIVE-SF] {\evaltable};
       \addlegendentry{MASSIVE SF}
       
       \addplot[very thick,draw=blue!100,dotted,mark=triangle*,mark options={scale=1.8,solid,fill=blue}] 
       table [x=size, y=MultiWoz-SF] {\evaltable};
       \addlegendentry{MultiWoz SF}
       \end{axis}

    \end{tikzpicture}
    \end{adjustbox}
    \caption{Performance of $\mbox{FLAN-T5}_{\scriptsize \mbox{~LoRA}}$ with various FLAN-T5 size.
    Solid-colored bars indicate adapters' training time for IC and striped bars for SF.
    }
    \label{fig:ablation-model-size}
\end{figure}
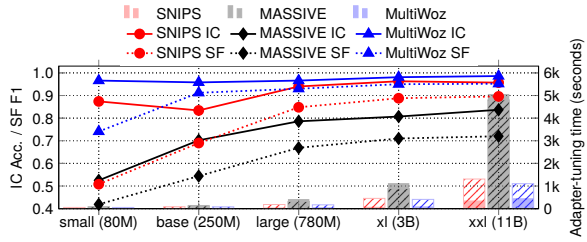

%% file: images/ablation_num_examples_per_label.tex
\pgfplotstableread{
num     SNIPS-IC    SNIPS-SF    MASSIVE-IC  MASSIVE-SF  MultiWoz-IC MultiWoz-SF
10      0.957       0.896       0.837       0.72        0.986       0.952             
20      0.967       0.922       0.847       0.743       0.987       0.952
50      0.966       0.937       0.84        0.743       0.986       0.957
100     0.966       0.93        0.862       0.74        0.983       0.96
all     0.967       0.948       0.871       0.797       0.989       0.962
}\evaltable

\pgfplotstableread{
num     SNIPS-IC    SNIPS-SF    MASSIVE-IC  MASSIVE-SF  MultiWoz-IC MultiWoz-SF
10      0.54        1.64        5.16        4.19        1.22        3.18   
20      1.07        3.25        10.21       8.27        2.04        6.20
50      2.68        8.36        24.59       20.58       4.53        15.49
100     5.35        16.64       44.6        41.18       8.37        31.19
all     100         100         100         100         100         100
}\datasizetable

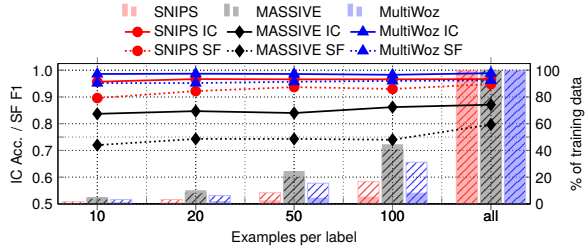
\begin{figure}[t]
    \centering
    \pgfplotsset{
       width=0.8\textwidth,
       height=0.2\textheight,
       grid=major,
       major grid style={dotted,draw=black!100},
       symbolic x coords={10,20,50,100,all},
       enlarge y limits={upper,value=0.05},
       legend style={
          draw=none,
          at={(0.5,1.0)},
          legend columns=3,
          legend cell align=left,
          anchor=south
          },
       }
    \begin{adjustbox}{width=0.48\textwidth}
    \begin{tikzpicture}
    \begin{axis}[
       xlabel={Examples per label},
       xtick pos=left,
       axis y line*=right,
       ybar,
       axis x line*=bottom,
       yticklabel pos=right,
       ylabel near ticks,
       ymajorgrids,
       bar width=0.5cm,
       ymin=0, ymax=100,
       ytick={0, 20, 40, 60, 80, 100},
       yticklabels={0, 20, 40, 60, 80, 100},
       ylabel style={align=center},
       ylabel={\% of training data},
       xtick=data,
       xticklabel style={
          inner sep=0pt,
          }
       ]
       \addplot[ybar,draw=none,ybar legend,fill=red!30] 
       table [x=num, y=SNIPS-IC] {\datasizetable};\label{SNIPS-IC}
       
       \addplot[ybar,draw=none,ybar legend,fill=black!30] 
       table [x=num, y=MASSIVE-IC] {\datasizetable};\label{MASSIVE-IC}
       
       \addplot[ybar,draw=none,ybar legend,fill=blue!30] 
       table [x=num, y=MultiWoz-IC] {\datasizetable};\label{MultiWoz-IC}
    \end{axis}
    \begin{axis}[
       ybar,
       bar width=0.5cm,
       ymin=0, ymax=100,
       xtick=data,
       yticklabels={,,},
       xticklabel style={
          inner sep=0pt,
          }
       ]
       \addplot[ybar,draw=red!30,ybar legend,pattern=north east lines,pattern color=red!80] 
       table [x=num, y=SNIPS-SF] {\datasizetable};\label{SNIPS-SF}
       
       \addplot[ybar,draw=black!30,ybar legend,pattern=north east lines,pattern color=black!80] 
       table [x=num, y=MASSIVE-SF] {\datasizetable};\label{MASSIVE-SF}
       
       \addplot[ybar,draw=blue!30,ybar legend,pattern=north east lines,pattern color=blue!80] 
       table [x=num, y=MultiWoz-SF] {\datasizetable};\label{MultiWoz-SF}
    \end{axis}
    \begin{axis}[
       axis y line*=left,
       axis x line*=bottom,
       xticklabels={},
       yticklabel pos=left,
       ylabel near ticks,
       ymin=0.5, ymax=1.0,
       ytick={0.5, 0.6, 0.7, 0.8, 0.9, 1.0},
       yticklabels={0.5, 0.6, 0.7, 0.8, 0.9, 1.0},
       ylabel={IC Acc. / SF F1},
       ]
       \addlegendimage{draw=red!30,/pgfplots/refstyle=SNIPS-IC}
       \addlegendentry{SNIPS}
       \addlegendimage{draw=black!30,/pgfplots/refstyle=MASSIVE-IC}
       \addlegendentry{MASSIVE}
       \addlegendimage{draw=blue!30,/pgfplots/refstyle=MultiWoz-IC}
       \addlegendentry{MultiWoz}
       
       \addplot[very thick,draw=red!100,mark=*,mark options={scale=1.5,solid,fill=red}] 
       table [x=num, y=SNIPS-IC] {\evaltable};
       \addlegendentry{SNIPS IC}
       
       \addplot[very thick,draw=black!100,mark=diamond*,mark options={scale=1.8,solid,fill=black}] 
       table [x=num, y=MASSIVE-IC] {\evaltable};
       \addlegendentry{MASSIVE IC}
       
       \addplot[very thick,draw=blue!100,mark=triangle*,mark options={scale=1.8,solid,fill=blue}] 
       table [x=num, y=MultiWoz-IC] {\evaltable};
       \addlegendentry{MultiWoz IC}
       
       \addplot[very thick,draw=red!100,dotted,mark=*,mark options={scale=1.5,solid,fill=red}] 
       table [x=num, y=SNIPS-SF] {\evaltable};
       \addlegendentry{SNIPS SF}
       
       \addplot[very thick,draw=black!100,dotted,mark=diamond*,mark options={scale=1.8,solid,fill=black}] 
       table [x=num, y=MASSIVE-SF] {\evaltable};
       \addlegendentry{MASSIVE SF}
       
       \addplot[very thick,draw=blue!100,dotted,mark=triangle*,mark options={scale=1.8,solid,fill=blue}] 
       table [x=num, y=MultiWoz-SF] {\evaltable};
       \addlegendentry{MultiWoz SF}
       \end{axis}
    \end{tikzpicture}
    \end{adjustbox}
    \caption{Performance of $\mbox{FLAN-T5-xxl}_{\scriptsize \mbox{~LoRA}}$ with various number of examples ($k$) per label.
    Solid-colored bars indicate \% of training data for IC and striped bars for SF.
    }
    \label{fig:ablation-num-examples-per-label}
\end{figure}

%% file: images/ablation_peft_techniques.tex
\pgfplotstableread{
peft        SNIPS-IC    SNIPS-SF    MASSIVE-IC  MASSIVE-SF  MultiWoz-IC MultiWoz-SF
Prefix-tuning      0.871       0.318       0.527       0.302       0.987       0.903            
Prompt-tuning      0.946       0.358       0.728       0.159       0.982       0.954
LoRA        0.966       0.860       0.872       0.732       0.986       0.965
IA3         0.957       0.839       0.650       0.526       0.987       0.965
}\evaltable

\pgfplotstableread{
peft        SNIPS-IC
Prefix-tuning      0.018    
Prompt-tuning      0.002    
LoRA        0.085    
IA3         0.007    
}\datasizetable

\begin{figure}[t]
    \centering
    \pgfplotsset{
       width=0.8\textwidth,
       height=0.2\textheight,
       grid=major,
       major grid style={dotted,draw=black!100},
       symbolic x coords={Prefix-tuning,Prompt-tuning,LoRA,IA3},
       enlarge y limits={upper,value=0.05},
       legend style={
          draw=none,
          at={(0.5,1.0)},
          legend columns=3,
          legend cell align=left,
          anchor=south
          },
       }
    \begin{adjustbox}{width=0.48\textwidth}
    \begin{tikzpicture}
    \begin{axis}[
       xlabel={PEFT technique},
       xtick pos=left,
       axis y line*=right,
       ybar,
       axis x line*=bottom,
       yticklabel pos=right,
       ylabel near ticks,
       ymajorgrids,
       bar width=0.5cm,
       ymin=0, ymax=0.100,
       ytick={0.001, 0.010, 0.100},
       yticklabels={0.001, 0.010, 0.100},
       ylabel style={align=center},
       ylabel={\% of model parameters},
       xtick=data,
       xticklabel style={
          inner sep=0pt,
          }
       ]
       \addplot[ybar,draw=none,ybar legend,fill=blue!30] 
       table [x=peft, y=SNIPS-IC] {\datasizetable};\label{PARAMS}
       
       
    \end{axis}
    \begin{axis}[
       axis y line*=left,
       axis x line*=bottom,
       xticklabels={},
       yticklabel pos=left,
       ylabel near ticks,
       ymin=0, ymax=1.0,
       ytick={0, 0.2, 0.4, 0.6, 0.8, 1.0},
       yticklabels={0, 0.2, 0.4, 0.6, 0.8, 1.0},
       ylabel={IC Acc. / SF F1},
       ]
       

       
       \addplot[very thick,draw=red!100,mark=*,mark options={scale=1.5,solid,fill=red}] 
       table [x=peft, y=SNIPS-IC] {\evaltable};
       \addlegendentry{SNIPS IC}
       
       \addplot[very thick,draw=black!100,mark=diamond*,mark options={scale=1.8,solid,fill=black}] 
       table [x=peft, y=MASSIVE-IC] {\evaltable};
       \addlegendentry{MASSIVE IC}
       
       \addplot[very thick,draw=blue!100,mark=triangle*,mark options={scale=1.8,solid,fill=blue}] 
       table [x=peft, y=MultiWoz-IC] {\evaltable};
       \addlegendentry{MultiWoz IC}
       
       \addplot[very thick,draw=red!100,dotted,mark=*,mark options={scale=1.5,solid,fill=red}] 
       table [x=peft, y=SNIPS-SF] {\evaltable};
       \addlegendentry{SNIPS SF}
       
       \addplot[very thick,draw=black!100,dotted,mark=diamond*,mark options={scale=1.8,solid,fill=black}] 
       table [x=peft, y=MASSIVE-SF] {\evaltable};
       \addlegendentry{MASSIVE SF}
       
       \addplot[very thick,draw=blue!100,dotted,mark=triangle*,mark options={scale=1.8,solid,fill=blue}] 
       table [x=peft, y=MultiWoz-SF] {\evaltable};
       \addlegendentry{MultiWoz SF}
       \end{axis}
    \end{tikzpicture}
    \end{adjustbox}
    \caption{Performance of FLAN-T5-xxl with different PEFT techniques. Bars indicate \% of model parameters trained during PEFT.
    }
    \label{fig:ablation-peft-techniques}
\end{figure}
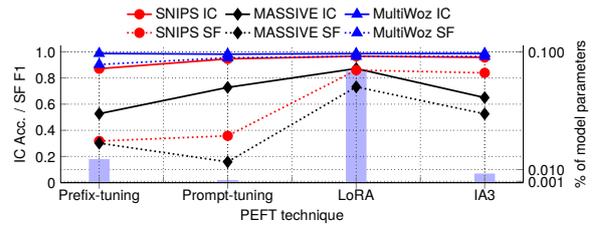

%% file: images/ablation_peft_generalization.tex
\pgfplotstableread{
peft        SNIPS-IC    SNIPS-SF    MASSIVE-IC  MASSIVE-SF  MultiWoz-IC     MultiWoz-SF
Prefix-tuning      0.046       0.316       0.025       0.0475       0              0.018            
Prompt-tuning      0.848       0.376       0.9615      0.3555       0.8145         0.124
LoRA        0.8385      0.539       0.9415      0.65         0.8335         0.5
IA3         0.822       0.5955      0.951       0.5035       0.8315         0.5305
}\evaltable

\pgfplotstableread{
peft        IC      SF
Prefix-tuning      0.024   0.127 
Prompt-tuning      0.875   0.285 
LoRA        0.871   0.563 
IA3         0.868   0.543 
}\datasizetable

\begin{figure}[t]
    \centering
    \pgfplotsset{
       width=0.8\textwidth,
       height=0.2\textheight,
       grid=major,
       major grid style={dotted,draw=black!100},
       symbolic x coords={Prefix-tuning,Prompt-tuning,LoRA,IA3},
       enlarge y limits={upper,value=0.05},
       legend style={
          draw=none,
          at={(0.5,1.0)},
          legend columns=3,
          legend cell align=left,
          anchor=south
          },
       }
    \begin{adjustbox}{width=0.48\textwidth}
    \begin{tikzpicture}
    \begin{axis}[
       xlabel={PEFT technique},
       xtick pos=left,
       axis y line*=left,
       ybar,
       axis x line*=bottom,
       yticklabel pos=left,
       ylabel near ticks,
       ymajorgrids,
       bar width=0.5cm,
       ymin=0, ymax=1.0,
       ytick={0, 0.2, 0.4, 0.6, 0.8, 1.0},
       yticklabels={0, 0.2, 0.4, 0.6, 0.8, 1.0},
       ylabel style={align=center},
       ylabel={IC Avg. Acc. / SF Avg. F1},
       xtick=data,
       xticklabel style={
          inner sep=0pt,
          }
       ]
       \addplot[ybar,draw=none,ybar legend,fill=red!30] 
       table [x=peft, y=IC] {\datasizetable};\label{AVG-IC}

       \addplot[ybar,draw=none,ybar legend,fill=blue!30] 
       table [x=peft, y=SF] {\datasizetable};\label{AVG-SF}      

       \addlegendimage{draw=red!30,/pgfplots/refstyle=AVG-IC}
       \addlegendentry{Average IC}   
        
       \addlegendimage{draw=blue!30,/pgfplots/refstyle=AVG-SF}
       \addlegendentry{Average SF}

    \end{axis}
    \end{tikzpicture}
    \end{adjustbox}
    \caption{Generalization of FLAN-T5-xxl with different PEFT techniques. Bars indicate average model performance across datasets.
    }
    \label{fig:ablation-peft-generalization}
\end{figure}
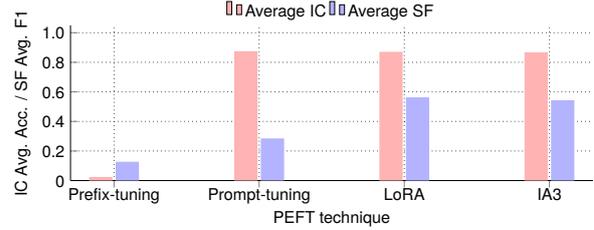

%% file: images/ablation-exposure-labels.tex
\begin{table}[t!] 
\centering
\begin{adjustbox}{width=0.48\textwidth}
\small
\begin{tabular}{l cc cc cc}
        \toprule
         & \multicolumn{2}{c}{\textbf{SNIPS}} & \multicolumn{2}{c}{\textbf{MASSIVE}} & 
         \multicolumn{2}{c}{\textbf{MultiWoZ}}\\
        \cmidrule(lr){2-3} \cmidrule(lr){4-5} \cmidrule(lr{0em}){6-7}
        & \textsc{test}$-$ & \textsc{test}$+$ & \textsc{test}$-$ & \textsc{test}$+$ & \textsc{test}$-$ & \textsc{test}$+$\\
        \midrule
        \shaderow \multicolumn{7}{l}{Intent Classification (Acc.)}\\
        \textsc{train}$-$  & 0.930 & 0.946 & 0.809 & 0.777 &  0.973 & \textbf{0.986} \\
        \textsc{train}$+$ & 0.136 & \textbf{0.962} & 0.062 & \textbf{0.825} &  0.124 & 0.979 \\
        \midrule
        \shaderow \multicolumn{7}{l}{Slot Filling (F1)}\\
        \textsc{train}$-$ & 0.002 & 0.143 & 0.033 & 0.207 &  0.161 & 0.389 \\
        \textsc{train}$+$ & 0.000 & \textbf{0.909} & 0.001 & \textbf{0.735} &  0.000 & \textbf{0.945} \\        
        \bottomrule
        \end{tabular}
    \end{adjustbox}
\caption{Effect of fine-tuning and evaluation with ($+$) and without ($-$) labels in the instructions. 
}
\label{tab:exposure_labels}    
\end{table}

%% file: images/ablation-multilinguality.tex
\begin{table}[!t] 
\centering 
\small
\begin{adjustbox}{width=0.48\textwidth}
\begin{tabular}{@{}lc@{\hspace{4pt}}cc@{\hspace{4pt}}cc@{\hspace{4pt}}cc@{\hspace{4pt}}cc@{\hspace{4pt}}c@{}}
\toprule
\multirow{3}{*}{\bf Model} & \multicolumn{10}{c}{\textbf{MASSIVE}} \\
 & \multicolumn{2}{c}{\bf en} & \multicolumn{2}{c}{\bf de} & \multicolumn{2}{c}{\bf fr} & \multicolumn{2}{c}{\bf it} & \multicolumn{2}{c}{\bf es} \\
\cmidrule(lr@{0em}){2-3} \cmidrule(lr@{0em}){4-5} \cmidrule(lr@{0em}){6-7} \cmidrule(lr@{0em}){8-9} \cmidrule(lr@{0em}){10-11}
 & IC & SF & IC & SF & IC & SF & IC & SF & IC & SF \\
 \midrule
 $\mbox{flan-t5-xxl}_{\scriptsize \mbox{~LoRA}}$ & \textbf{.833} & \textbf{.735} & \textbf{.783} & .669 & .770 & .614 & .745 & .573 & .733 & .560 \\
 \midrule
 \shaderow \multicolumn{11}{@{}l@{}}{Multilingual LLMs} \\
 $\mbox{mt5-xxl}_{\scriptsize \mbox{~LoRA}}$ & .795 & .674 & .775 & .653 & \textbf{.784} & .671 & .773 & .663 & .773 & .649 \\
 $\mbox{mt0-xxl}_{\scriptsize \mbox{~LoRA}}$ & .804 & .689 & .778 & .662 & \textbf{.784} & .660 & \textbf{.797} & .658 & \textbf{.784} & .631 \\
 $\mbox{mt0-xxl-mt}_{\scriptsize \mbox{~LoRA}}$ & .814 & .700 & .781 & \textbf{.689} & .685 & \textbf{.679} & .751 & \textbf{.679} & .768 & \textbf{.670} \\
\bottomrule
\end{tabular}
\end{adjustbox}
\caption{Performance on languages other than English in terms of intent accuracy (IC) and slot filling F1 (SF).}
\label{tab:multilinguality}
\end{table}

%% file: tables/discussion.tex
\newcolumntype{u}{>{\hsize=.8\hsize}X}
\newcolumntype{s}{>{\hsize=.1\hsize}X}
\newcolumntype{b}{>{\hsize=1.9\hsize}X}
\newcolumntype{m}{>{\hsize=1.2\hsize}X}

\begin{table*}[th!]
\centering
\scriptsize
\begin{adjustbox}{width=\textwidth}
    \begin{tabularx}{\textwidth}{s u b mum@{}}
    \toprule
        \textbf{ID} & \multirow{2}{=}{\textbf{Problem Category}}& \textbf{User utterance} & \textbf{Expected Label(s)} & \multicolumn{2}{c}{\textbf{LLM Response}} \\
        \cmidrule(l{0em}r{0em}){5-6}
         &  &  &  & {\bf ILLUMINER} & {\bf $\mbox{GPT3.5}_{\tiny \mbox{~few-shot}}$} \\
    \midrule
    1 & Ambiguous User Utterances
        & ``pink is all we need'' & $l$: change light color & \textcolor{red}{express liking music} & \textcolor{red}{be quirky} \\
    \midrule
    2 & Entity Disambiguation
        & ``what's the time in australia'' & $s$: place-name: australia & \textcolor{red}{time-zone}: australia & \textcolor{red}{time-zone}: australia \\
    \midrule
    3 & Missing Context
        & ``remind me to do something then'' & $l$: set a calendar event & \textcolor{red}{set an alarm}  & \textcolor{red}{set an alarm}\\
    \midrule
    4 & Highly correlated labels 
        & ``put lindsey cardinale into my hillary clinton s women s history month playlist'' & $s$: artist: lindsey cardinale & artist: lindsey cardinale & \textcolor{red}{entity-name}: lindsey cardinale \\  
    \midrule
    5 & Hallucinations
        & ``turn my morning alarm on'' & $l$: set an alarm & \textcolor{red}{turn an alarm on} & set an alarm \\
    \cmidrule(lr{0em}){3-6}
    6 &  & ``play it again please'' & $s: \emptyset$ & $\emptyset$ & \textcolor{red}{player-setting}: \textcolor{red}{repeat} \\
    \bottomrule
\end{tabularx}
\end{adjustbox}
\caption{Problem categories with exemplars. $l$ and $s$ denote expected intents and slots.
We report LLM predictions by ILLUMINER ($\mbox{flan-t5-xxl}_{\scriptsize \mbox{~LoRA}}$) and $\mbox{GPT3.5}_{\tiny \mbox{~few-shot}}$. Erroneous predictions are marked red.}
\label{tab:discussion}    
\end{table*}